\documentclass[letterpaper, 10 pt, conference]{ieeeconf}  %

\IEEEoverridecommandlockouts                              %
\usepackage{amsmath,amssymb,amsfonts}
\usepackage{algorithmic}
\usepackage{graphicx}
\usepackage{textcomp}
\usepackage{biblatex}
\usepackage{xcolor}
\usepackage{flushend}
\def\BibTeX{{\rm B\kern-.05em{\sc i\kern-.025em b}\kern-.08em
    T\kern-.1667em\lower.7ex\hbox{E}\kern-.125emX}}
\begin{document}

\title{\LARGE \bf
Dense Crowds Detection and Surveillance with Drones using Density Maps*
}

\author{Gonzalez-Trejo, Javier$^{1}$ \and Mercado-Ravell, Diego.$^{2}$%
\thanks{This work was supported by the Mexican National Council of Science and Technology CONACYT.}
\thanks{$^{1}$J. Gonzalez-Trejo is with the Center for Research in Mathematics CIMAT-Zac, Unit Zacatecas, Mexico, 
        {\tt\small email: javier.gonzalez@cimat.mx}}%
\thanks{$^{2}$D. Mercado-Ravell is with Cátedras CONACYT at the Center for Research in Mathematics CIMAT-Zac, Unit Zacatecas, Mexico, (corresponding author) phone: +52-449-428-4800;
        {\tt\small email: diego.mercado@cimat.mx}}%
}

\maketitle
\thispagestyle{empty}
\pagestyle{empty}

\begin{abstract}
Detecting and Counting people in a human crowd from a moving drone present challenging problems that arise from the constant changing in the image perspective and camera angle.
In this paper, we test two different state-of-the-art approaches, density map generation with VGG19 trained with the Bayes loss function and detect-then-count with Faster RCNN with ResNet50-FPN as backbone, in order to compare their precision for counting and detecting people in different real scenarios taken from a drone flight.
We show empirically that both proposed methodologies perform especially well for detecting and counting people in sparse crowds when the drone is near the ground. Nevertheless, VGG19 provides better precision on both tasks while also being lighter than Faster RCNN. Furthermore, VGG19 outperforms Faster RCNN when dealing with dense crowds, proving to be more robust to scale variations and strong occlusions, being more suitable for surveillance applications using drones.
\end{abstract}

\section{Introduction}

\begin{figure}[t!]
\centerline{\includegraphics[width=0.5\textwidth]{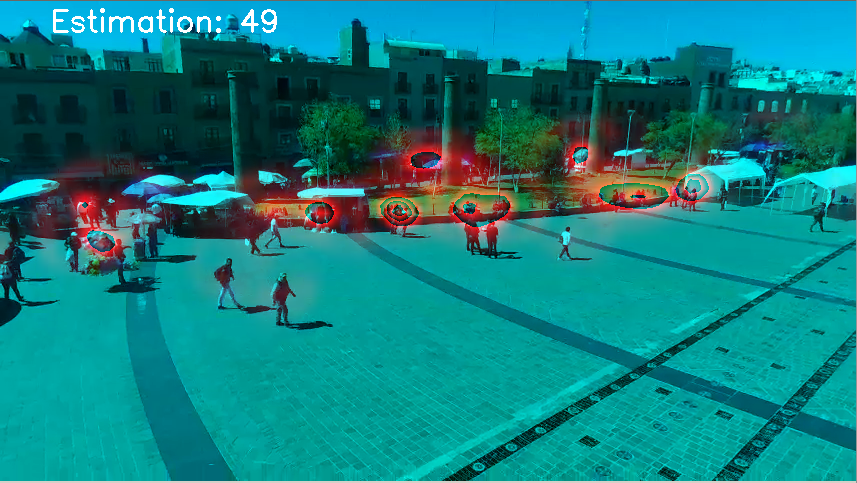}}
\caption{Density map generated by VGG19 taken from a drone in a large public square. The ground truth count is 48, while the estimated is 49 persons.}
\label{vgg_bicentenary}
\end{figure}

The use of drones for the tasks of human crowd detection and counting has taken relevance by the fact that the drone can move freely, thus it is easier to monitor big crowds using less cameras since,  in general, one drone might be sufficient to provide a good estimate on how dense the crowd is. In addition, it is capable to detect and track without lousing the crowd or person target \cite{sirmacek11_autom},\cite{songchenchen19_method_based_multi_featur_detec},\cite{Pareek_2019},\cite{shao18_using_multi_scale_infrar_optic}, something that can hardly be accomplished with only stationary cameras.
\\However, the use of crowd density estimation is not only limited to perform surveillance. Another relevant use case where crowd detection is needed, is in drone's autonomous landing in crowded places \cite{8081306},\cite{8657776},\cite{kuchhold18_scale_adapt_real_time_crowd},\cite{Liu_2019_CVPR},\cite{8868691}. The objective there is to infer the crowd density and land in place, where ideally there is not a single persons on a safe radius. Nevertheless this kind of tasks usually requires to overestimate the detection for safety reasons.\\
In the classical approach, detecting-then-counting is used to perform the count, from either images from a static camera or airborne devices \cite{ge09_marked}. However this approaches are susceptible to body occlusions or few pixels per persons, considering that they try to find the whole body of the person. Thus, making them only useful for low density crowds.\\
Moving forward, researchers found that, since the head is the most visible part of a person in a crowd, there was no need to detect the full body on the image and it was enough to detect the heads in a crowded pattern. This approach generates density maps as shown in Figure \ref{vgg_bicentenary}, which are represented as a heat map of the estimated number of human heads founded in an image, using low level futures \cite{chan08_privac} like Maximum Excess over SubArrays (MESA) \cite{NIPS2010_4043}, Features from Accelerated Segment Test (FAST) \cite{8451289} or features dependent on the movement \cite{rabaudil_count_crowd_movin_objec}, \cite{loy13_from_semi_trans_count_crowd}. Then again, they were not robust enough and tend to detect too many false positives, or depended on a fixed perspective in order to work.
\\In the context of drones, changes in the perspective causes distortions and uneven human sizes by the constant drone's movement. To solve this, many approaches from the perspective of the algorithm using deep learning were proposed.
Some of these algorithms introduce multi-column architectures that take into account the different sizes between heads \cite{zhang16_singl_image_crowd_count_multi}, generating a scale map taken from the image in order to provide the different head sizes in the scene \cite{basalamah19_scale_driven_convol_neural_networ}. Other works use the Deep Neural Network only as a head detector, in order to provide information on where to perform the count \cite{shami19_peopl_count_dense_crowd_images}.
Since crowd detection and counting from the drone is still an uncharted area, there are only few proposals that take into account information from the drone. The most relevant ones use information of the pitch angle and altitude from the drone to produce a perspective map, that is fed into a context aware architecture, which then produces a density map that can be used, for example, in landing the drone away from the crowd \cite{Liu_2019_CVPR}, \cite{liu19_geomet_physic_const_drone_based}. All of this approaches use the ground truth density maps as learning targets. The heads annotations from which these density maps are generated, present small but substantial errors that might force any deep learning architecture to learn incorrect features from the images \cite{ma2019bayesian}.\\
In this paper, we capture video streams from a drone under different real scenarios. Then, such video streams are processed in a ground computer using ROS and Pytorch. There, we utilize the VGG19 architecture with pre-trained weights on the UCF-QNRF database using the Bayes loss function \cite{ma2019bayesian} to generate density maps of each frame. Furthermore, we compare it against Faster RCNN with ResNet50-FPN \cite{lin2016feature} as backbone, using the strategy of detecting persons proposing a bounding box where it finds the targets, and then counts them. We found that while both performed relatively well on all scenarios, VGG19 with the Bayes loss function is better suited for crowd counting and detection in drones.
\\The organization of this paper is as follows. In Section \ref{sec-previous}, we discuss related work on crowd detection and counting with focus on drones. Then, we briefly describe  the structure of both VGG19 and Faster RCNN in Section \ref{sec-method}. Also, we evaluate and discuss the experimental results in different real scenarios, while comparing both methodologies in Section \ref{sec-experiment}. Finally we present our conclusions and future work in Section \ref{sec-conclusions}.

\section{Previous work}
\label{sec-previous}
Initial work in the field of crowd detection and counting focused primarily on regression methods in images from static cameras used for surveillance \cite{rabaudil_count_crowd_movin_objec}, \cite{chan08_privac}, \cite{loy13_from_semi_trans_count_crowd}, \cite{ge09_marked}. More in specific, works like \cite{rabaudil_count_crowd_movin_objec}, \cite{loy13_from_semi_trans_count_crowd}, \cite{ge09_marked} used some sort of feature extractor over a singular image, or continuous videos, to perform the count. These methods suffer from a series of pitfalls, for example not accounting for the image perspective, that is, expecting people's bodies or heads to be the same size across the image. The features extraction itself, that only searches for changes in the color like FAST, or movement like Kanade–Lucas–Tomasi feature tracker, are not robust enough at generalizing for images of dense crowds \cite{kang2018beyond}. The latter issue, is still suffered by methods like \cite{chan08_privac}, which despite solving the perspective problem by manually calculating it from the image, annotating the ROI (region of interest) for each surveillance camera.
\\More recent papers have identified the generation of density maps as a better tool to get count and position data in crowded scenes \cite{kang2018beyond}. In \cite{sirmacek12_kalman_filter_based_featur_analy} and \cite{sirmacek11_autom}, authors extracted FAST features from videos taken from an airborne camera above the crowd, provided as a Probability Density Function (PDF). Since the distance from the crowd with respect of the camera is constant, they employed a Kalman filter to also track the person's movement direction. Still, as previous mentioned works, this approach cannot generalise well in any other scenario, as it does not take into account perspective variations which are frequent when working with moving cameras, as a drone.
\\In more recent years, the use of DNN for crowd analysis in general, has taken more relevance, but very few works with focus on drones for crowd surveillance are to be found. Works that first used DNNs utilized the detect-then-count approach. For example, in \cite{wang16_skyey}, authors used Yolo (You Only Look Once) V3 for the detection of vehicle crowds. Since Yolo V3 is a heavy weight deep learning architecture, they proposed a model on which they used the drone as a mobile camera and the actual inferences are done in the cloud. For applications like vehicle crowds, this solution has his advantages, since the occlusion is less severe than in human crowds.
\\At current time, density map generators using DNN are the most popular methods for crowd counting, detection and tracking \cite{kang2018beyond}. In drones, this methods have been used for assistance in autonomous landing \cite{liu19_geomet_physic_const_drone_based}, \cite{8081306}, \cite{tzelepi19_graph_embed_convol_neural_networ}. In \cite{tzelepi19_graph_embed_convol_neural_networ}, authors used a lightweight Multitask DNN. This approach, does not look to individually detect and track each person in the crowd, since it is preferred to overestimate the density map, in order to prevent the drone  from landing near a group of persons. However, overestimating the detection is undesired for surveillance purposes.
\\Since the drone provides data of the pitch and the altitude \cite{liu19_geomet_physic_const_drone_based}, uses that information to provide an extra channel called the perspective map, to be fed into their DNN, from which the density map generated is drawn into the heads plane, it is, where the heads are in the real world, and not in the image plane. This strategy helps to prevent underestimation of near crowded scenes to help the drone to decide where to land without risking to hurt people.
\\ All this approaches deal with the problems of occlusions, scale variations and context diversity. Starting from the architectures that generate the density maps, which do not take into account that the datasets used for training are not correctly labeled in the first place, mainly due to the difficulty of precisely localizing the head center \cite{ma2019bayesian}. A direct consequence, specially when using multi-column strategies \cite{zhang16_singl_image_crowd_count_multi}, scale aware and context aware architectures slow down the crowd detection and counting tasks. In our work, we use VGG19 trained with the Bayes loss function to precisely solve this problem from the pre-training step, without forcing the creation of more complex architectures. As a direct consequence, future works, using the Bayes loss function could potentially result in the creation of lightweight architectures, with less than 1 million parameters, robust enough to deal in real time with perspective and scale changes from a moving video stream coming from a flying drone.

\section{Methodology}
\label{sec-method}
In this paper we propose two different approaches to accomplish crowd detection and counting, VGG19 with the Bayes Loss and Faster RCNN with ResNet5-FPN as backbone.

\subsection{Faster RCNN with ResNet50}

Faster RCNN with ResNet50-FPN stands for Faster Region-based Convolutional Neural Network (Faster RCNN) with Residual Network (ResNet) of 50 layers using Feature Pyramidal Networks (FPN).
Faster RCNN acts as our baseline for comparison in the crowd detection and counting tasks. It has three steps, the convolutional layers with the FPN, to extract the feature maps, the region proposal network, and the header which returns both the classification of the objects found and the coordinates of the bounding boxes, where the objects reside.

\subsubsection{Convolutional layers with FPN}
For this work, ResNet50 was proposed as the backend in Faster RCNN, where, the fully connected layers were removed from ResNet50. The output feature map $P_1$ is fed to the Region Proposal Network \cite{DBLP:journals/corr/RenHG015} and 4 additional feature maps $P_k$ where $k\in{2,3,4,5}$, and the higher the number $k$ is, the more semantic information is present in the feature map, at the cost of losing spatial information \cite{lin2016feature}.
\subsubsection{Region Proposal Network (RPN)}
Here, from the feature maps, ROIs are extracted and then each ROI's height $h$ and width $w$ are fed to the Equation (\ref{eq:roi}), from which a feature map $P_k$ is selected to be segmented using that ROI and then be fed into the ROI Pooling. The equation to select the feture map is as follows \cite{lin2016feature}:
\begin{equation}\label{eq:roi}
    k = \left\lfloor k_0 + \log_2(\sqrt{wh}/224) \right\rfloor
\end{equation}
where $k_0$ is the expected number associated with the feature map, here set to 4.
\subsubsection{ROI Pooling}
Lastly, the segmented feature maps are processed for classification and generation of the final bounding box. For this work, we only use the ``person" classification out of the numerous classes that COCO provides.

\subsection{VGG19 with Bayes Loss}
\begin{table}[b]
    \centering
    \caption{Structure of the VGG19 with regression header.}
    \begin{tabular}{|c|c|c|}
    \hline
    \textbf{Operation}     & \textbf{Kernel size} & \textbf{Output dimensions} \\
    \hline
    Conv2D & 3x3 & 64\\
    Conv2D & 3x3 & 64\\
    MaxPool2D & 2x2 & 64\\
    Conv2D & 3x3 & 128\\
    Conv2D & 3x3 & 128\\
    MaxPool2D & 2x2 & 128\\
    Conv2D & 3x3 & 256\\
    Conv2D & 3x3 & 256\\
    Conv2D & 3x3 & 256\\
    Conv2D & 3x3 & 256\\
    MaxPool2D & 2x2 & 256\\
    Conv2D & 3x3 & 512\\
    Conv2D & 3x3 & 512\\
    Conv2D & 3x3 & 512\\
    Conv2D & 3x3 & 512\\
    MaxPool2D & 2x2 & 512\\
    Conv2D & 3x3 & 512\\
    Conv2D & 3x3 & 512\\
    Conv2D & 3x3 & 512\\
    Conv2D & 3x3 & 512\\
    Conv2D & 3x3 & 256\\
    Conv2D & 3x3 & 128\\
    Conv2D & 1x1 & 1\\
    \hline
    \end{tabular}

    \label{tab:vgg19}
\end{table}
The VGG19 architecture, as described in Table \ref{tab:vgg19}, is used thanks to its great transfer learning capabilities. The main difference with the original implementation is the removal of the fully connected layers along with the last max pooling operation. Furthermore, it is attached to a regression header in order to generate the density map from the features that the VGG19 backend extracted from the image \cite{ma2019bayesian}.

\subsubsection{Ground truth density map generation}
Each image in a dataset that is used to count or detect crowds, is composed by the image where each person has a pixel annotation in the middle of his head. Since this annotated point is sparse and does not represent the person's head size, a 2D Gaussian distribution is used to blur out the point over an area, defined by the covariance matrix of the distribution. The 2D Gaussian distribution is evaluated for all $M$ pixels $\mathrm{x}_m$ in Equation \ref{eq:0} as follows \cite{ma2019bayesian}:
\begin{equation}\label{eq:0}
    D^{gt}(\mathrm{x}_m) =\sum^N_{n=1}\mathcal{N}(\mathrm{x}_m;\mathrm{z}_n, \sigma^{2}1_{2x2})
\end{equation}
where $D^{gt}(\mathrm{x}_m)$ is the 2D Gaussian distribution evaluated in the pixel $\mathrm{x}_m$, describing how much that pixel $\mathrm{x}_m$ accumulates for the total persons count $N$, with the mean defined at the annotated point location $\mathrm{z}_n$ and an isotropic covariance matrix $\sigma^{2}1_{2x2}$. The weights were trained with density maps of variance $\sigma=8$.

\subsubsection{Loss function}
Since, as stated in the introduction, the annotated point $y_n$ are prone to be miss-placed, the results of the density maps are used as likelihoods of point $\mathrm{x}_m$ given the annotated point $y_n$:
\begin{equation}\label{eq:1}
    p(\mathrm{x}_m|y_n) = \mathcal{N}(\mathrm{x}_m;\mathrm{z}_n, \sigma^{2}1_{2x2})
\end{equation}
to define the loss function, we need an a-posteriori probability of $\mathrm{x}_m$ given $y_n$ which is given by the Bayes rule, which gives its name to the loss function, as follows \cite{ma2019bayesian}:
\begin{equation}\label{eq:2}
    p(y_n|\mathrm{x}_m) = \frac{p(\mathrm{x}_m|y_n)}{\sum^N_{n=1}p(\mathrm{x}_m|y_n)}
\end{equation}
assuming that the probability $p(y_n)$ of finding an annotated point in the image is equal to $1/N$. With Equation (\ref{eq:2}) we now can obtain the expected total count $E[c_n]$ associated with $y_n$ from all the estimated values $D^{est}(\mathrm{x}_m)$ as 
\begin{equation}\label{eq:3}
    E[c_n] = \sum^M_{m=1}p(y_n|\mathrm{x}_m) D^{est}(\mathrm{x}_m)
\end{equation}
the expected count $E[c_n]$ for an annotation point $y_n$ will be enough to construct the Bayes loss function, but as it is explained in \cite{ma2019bayesian}, to provide a more robust loss function, the modeling of an annotation belonging to the background $y_0$ is needed. For the position of the background, an annotation point $\mathrm{z}^m_0$ is defined as:
\begin{equation}\label{eq:4}
    \mathrm{z}_0^m = \mathrm{z}_n^{m} + d\frac{\mathrm{x}_m - \mathrm{z}^m_n}{||\mathrm{x}_m - \mathrm{z}^m_n  ||_2} 
\end{equation}
were each background annotation $\mathrm{z}^m_0$ is defined by both $\mathrm{x}_m$ and it's nearest annotation point $\mathrm{z}^m_n$. The distance ratio $d$ defines by how much the $\mathrm{z}^m_0$ will be far from $\mathrm{z}^m_n$. In this case $d$ was set to 0.15. Once the background annotation is obtained, we proceed to define the expected total count $c_0$, for the annotated background point $y_0$ as with $y_n$

\begin{equation}\label{eq:5}
    p(\mathrm{x}_m|y_0) = \mathcal{N}(\mathrm{x}_m;\mathrm{z}_0^m, \sigma^{2}1_{2x2})
\end{equation}
\begin{equation}\label{eq:6}
    p(y_0|\mathrm{x}_m) = \frac{p(\mathrm{x}_m|y_n)}{\sum^N_{n=1}p(\mathrm{x}_m|y_n) + p(\mathrm{x}_m|y_0)}
\end{equation}

\begin{equation}\label{eq:7}
    E[c_0] = \sum^M_{m=1}p(y_0|\mathrm{x}_m) D^{est}(\mathrm{x}_m)
\end{equation}
assuming, that the probability $p(y_0)$ of finding a background point is equal to the probability of finding an annotated head $p(y_n)$, now set to $1/(N+1)$, and taking into account the new annotated point.\\
Now we combine both equations (\ref{eq:7}) and (\ref{eq:3}) into the loss function $\mathcal{L}^{loss}$ \cite{ma2019bayesian}:
\begin{equation}\label{eq:8}
    \mathcal{L}^{loss}=\sum_{n=1}^N\mathcal{F}(1-E[c_n]) + \mathcal{F}(0-E[c_0])
\end{equation}
knowing that $\mathcal{F}$ is the distance function, in this case the first norm. The
expected count value for each annotated head point is 1 and the expected count value in the background is 0.

\section{Experiments}
\label{sec-experiment}
\subsection{Experimental setup}
\begin{figure}[t]
\centerline{\includegraphics[width=0.5\textwidth]{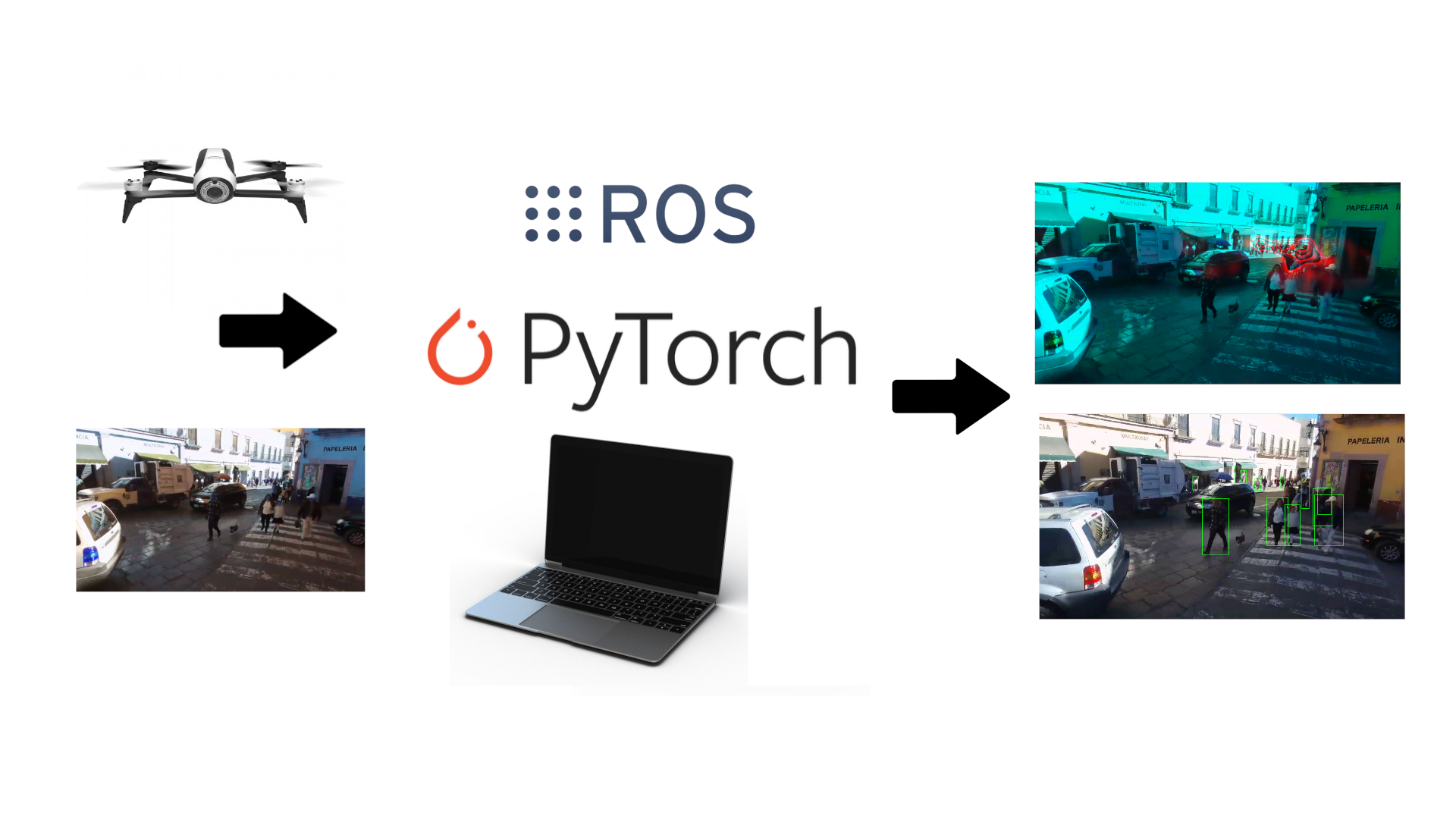}}
\caption{We utilize the drone Bebop Parrot 2 to obtain the video stream back to the ground laptop computer using ROS. Utilizing the PyTorch framework, we process the video stream in 2 ROS nodes, one with FasterRCNN ResNet50 and the other one with VGG19, trained with the Bayes Loss to generate density maps.}
\label{info}
\end{figure}
We tested Faster RCNN ResNet50 and VGG19 trained with Bayes loss on real scenarios using video streams of resolution 
$480\times856$ pixels in 30 frames per second, gathered from a drone. Just as we showed in Figure \ref{info}, we setup a ground laptop computer with an Intel Core i7 paired with a Nvidia GTX 1050 with 4 GB of video RAM and 8 GB of regular RAM. We directed the drone over the scene while capturing the video stream to post-process it using the nodes that contained the DNN under evaluation.
To perform the counting, we accumulated all the classes `person' that Faster RCNN classified, and accumulated the density map output by the VGG19. We compared the accumulation of the density map and the total detections in the most relevant frame of each scenario. It is important to notice that this experimental setup is ready to deal with the problem in real-time. However, the considered methods may take up to 4 seconds to run under our current hardware.\\
As a note, all the images produced with VGG19 seem blue because we substituted the density map instead of the red channel, just for a clearer depiction.
\subsection{Real Scenarios}
We recorded 6 different scenarios, from low to moderately crowded. A compilation can be found in the following link \url{https://youtu.be/C5KGAajiJ50}. Now we describe each of these real scenarios:
\begin{itemize}
    \item \textbf{Garden: } Here, the camera is near at the same altitude of the person's height. It presents more urban scenarios with cars passing by. The crowd level is low to moderated.
    Same as Fountain but with better light.
    \item \textbf{Small square: }Here the crowd levels are low with some stands, balloons and trees. It stands as the most difficult scenario for both approaches.
    \item \textbf{Large public square: }It was recorded in an open space near a bus station, with hills as background. The most challenging features come precisely from the hills, trees and some stands where human and round figures appear. The scene is mostly low crowded.
    \item \textbf{Public University : }The drone hovers in front of a staircase in the middle of a university campus. It presents a rural background, with hills, trees and rocks. This rural-like scenario presents challenging patterns for the VGG19 and all density map generators.
   \item \textbf{Fountain: } Similar to garden but with worst lighting.
\end{itemize}

Following, we present the results obtained from the scenarios described before.

\subsubsection{Garden}
\begin{figure}[t]
\centerline{\includegraphics[width=0.5\textwidth]{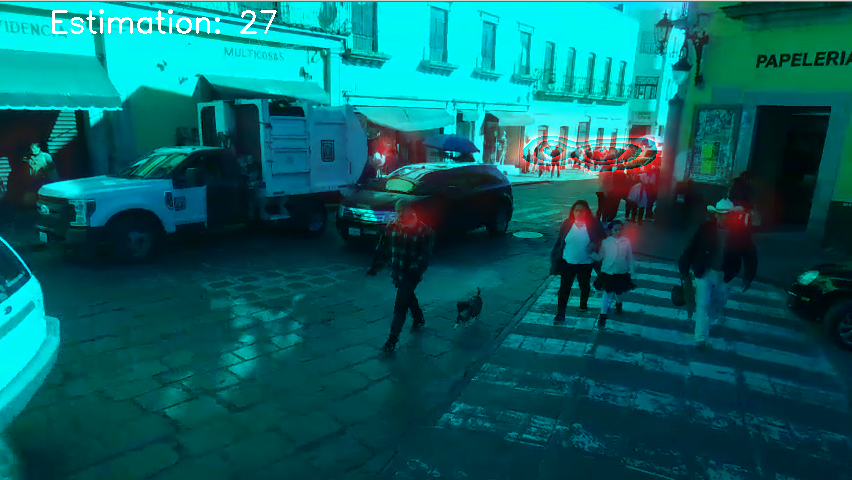}}
\caption{Relevant frame in the garden scenario using VGG19 as detector. The estimated count is 27 against the ground truth count of 25 persons. VGG19 is able to detect the persons far away from the camera without problems.}
\label{vgg_garden}
\end{figure}

\begin{figure}[t]
\centerline{\includegraphics[width=0.5\textwidth]{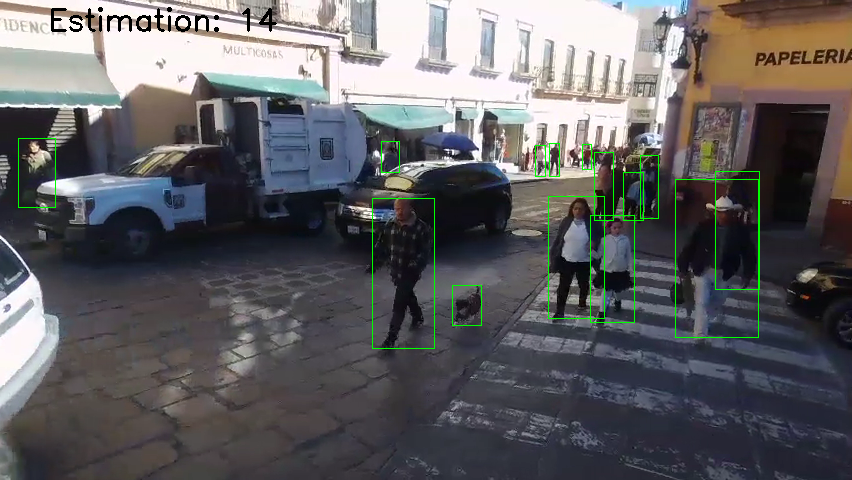}}
\caption{Relevant frame in the garden scenario using Faster RCNN as detector. The estimated count is 14 persons against the ground truth count of 25 persons. Only few people that are far away from the camera are detected. The dog is incorrectly detected as a person, therefor only 13 persons were detected correctly.}
\label{resnet_garden}
\end{figure}

For most of the video, both VGG19 and Faster RCNN perform equivalently, yet differences arise when heavy occlusions and few pixels per persons are present on the image. As we show in Figures \ref{vgg_garden} and \ref{resnet_garden}, both methods can detect and classify correctly as ``person"s the 6 targets nearer to the camera, only having the error of classifying the dog as a person by Faster RCNN. People far away from the camera are not recognized by Faster RCNN, and even a human would have struggled to identify them and count them correctly. Nevertheless, VGG19 succeeds at localizing them and only fails by two persons on the counting task.

\subsubsection{Small square}
\begin{figure}[t]
\centerline{\includegraphics[width=0.5\textwidth]{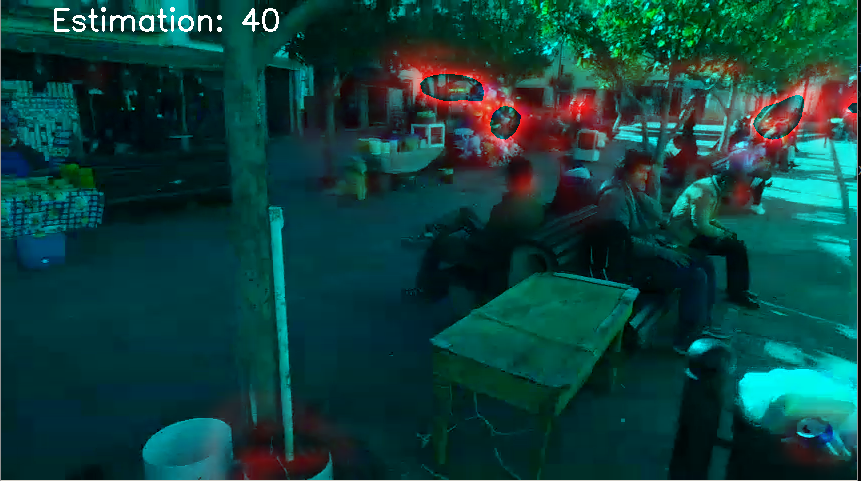}}
\caption{Relevant frame in a small public square using VGG19 as the detector. The estimated count is 40 persons against the ground truth count of 17. VGG19 fails at counting balloons as a group of persons, probably  because of the round and repetitive pattern of the balloons.}
\label{vgg_genaro}
\end{figure}

\begin{figure}[t]
\centerline{\includegraphics[width=0.5\textwidth]{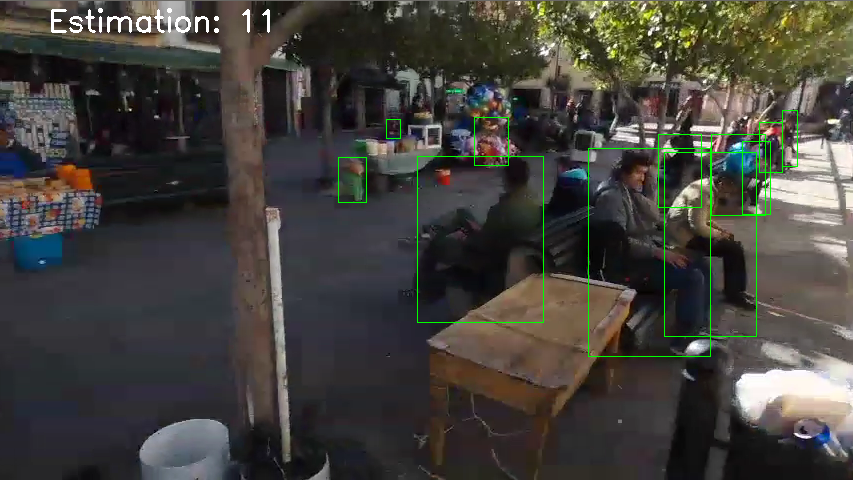}}
\caption{Relevant frame in the small square scenario using Faster RCNN as the detector. The estimated count is 11 persons against the ground truth count of 17 persons. Even though Faster RCNN detects both the toy stand and stools as persons, since the false positive detection are only as single persons, the estimated count is more reasonable compared to the ground truth.}
\label{resnet_genaro}
\end{figure}

The frame chosen from this video is interesting since it has heavily occluded persons that also are far away from the camera, while presenting rich patterns coming from the trees and some of the stands. As we show in Figure \ref{vgg_genaro}, VGG19, and in general density maps generators, tend to fail when they find round repetitive patterns that look like groups of person's heads. Even though VGG19 correctly estimated the density from persons far away of the camera, detecting the toy stand as a group of persons increased by a significant amount the estimated count, making it unreliable for this kind of scenarios.\\ Faster RCNN as shown in Figure\ref{resnet_genaro}, provides a more reasonable estimate, even though it only counts the persons near the camera.

\subsubsection{Large public square}

\begin{figure}[t]
\centerline{\includegraphics[width=0.5\textwidth]{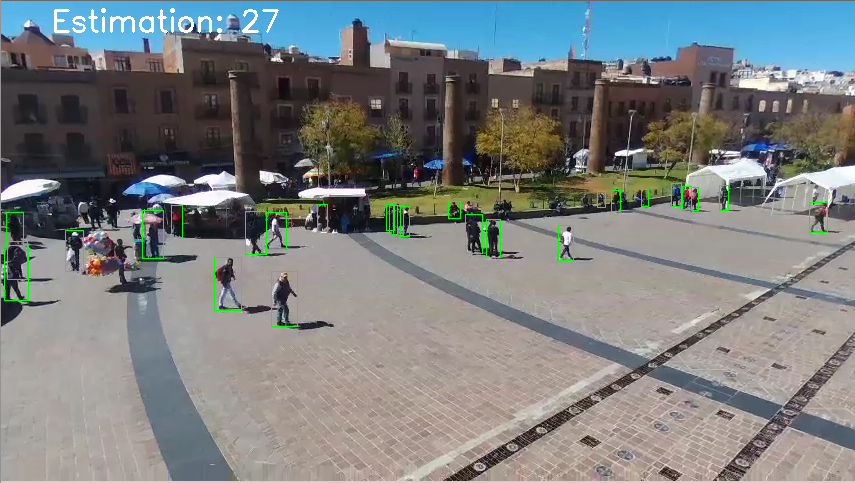}}
\caption{Relevant frame in the large public square scenario, using Faster RCNN as the detector. The estimated count is 27 persons against the ground truth count of 49 persons. Here, Faster RCNN fails at counting people siting near the yard.}
\label{resnet_bicentenary}
\end{figure}

Here, the drone flights at around 4 meters over the ground, making the pixel count per person small. Both VGG19 and Faster RCNN perform similarly for the detecting and counting tasks, when the persons have their body fully visible and are near the camera. For Faster RCNN, as shown in the Figure \ref{vgg_bicentenary} compared to Figures \ref{resnet_genaro} and \ref{resnet_garden}, it better detects persons with fewer pixels, but fails more often to count sitting persons compared with Figure \ref{resnet_genaro}. VGG19 performed near perfect for this frame, since it was able to estimate 47 persons out of 48, yet sometimes when hills with rich patterns show ups in the background, VGG19 miss-classify them as dense crowds.
\subsubsection{Public university}
\begin{figure}[t]
\centerline{\includegraphics[width=0.5\textwidth]{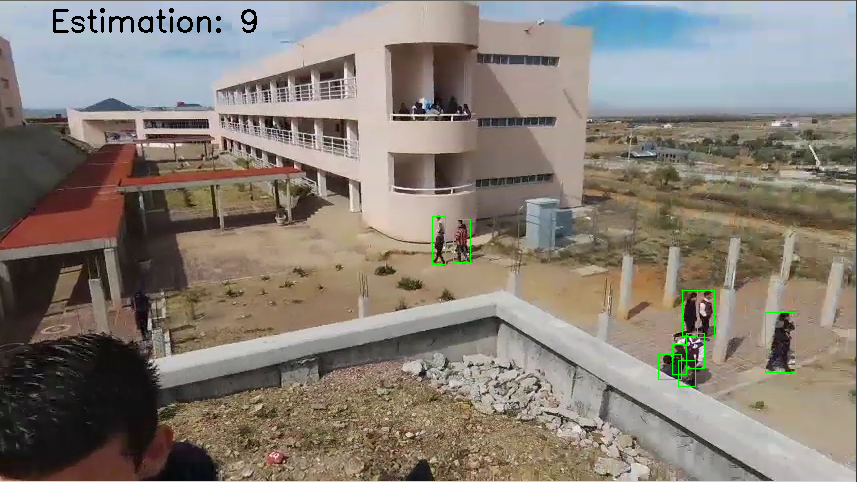}}
\caption{Relevant frame in the public university scenario, using Faster RCNN as the detector. The estimated count is 9 persons against the ground truth count of 20 persons. The persons on the balcony are not detected and false positives show near the stairs on the right of the image.}
\label{resnet_poli}
\end{figure}
\begin{figure}[t]
\centerline{\includegraphics[width=0.5\textwidth]{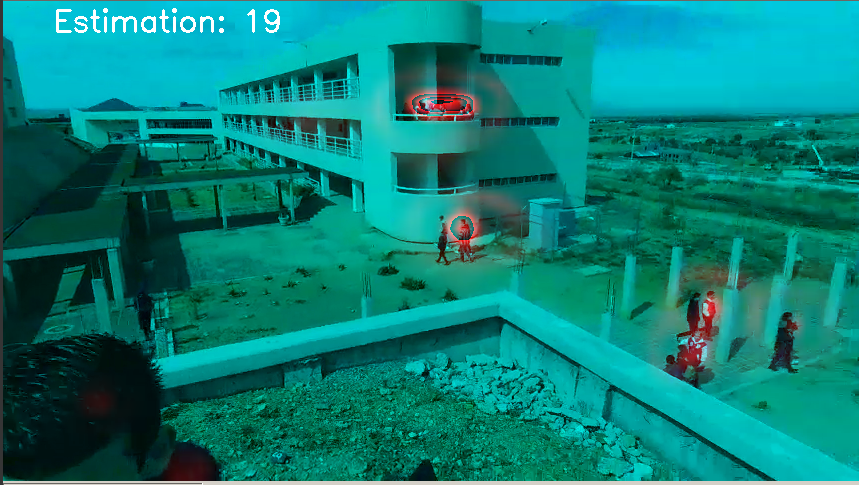}}
\caption{Relevant frame in the public university scenario, using VGG19 as the detector. The estimated count is 19 persons against the ground truth count of 20 persons. The only persons not detected by VGG19 are the ones that merge with the shadows, behind the person's head in the left of the image.}
\label{vgg_poli}
\end{figure}
As we can see in Figure \ref{resnet_poli}, Faster RCNN could not capture the persons in the balcony and counted more than once some persons. Here VGG19, and for the most part of the video, has the most approximate estimated count to the ground truth, despite of the hills and rocks presented in the scenery.
\subsubsection{Fountain}
\begin{figure}[t]
\centerline{\includegraphics[width=0.5\textwidth]{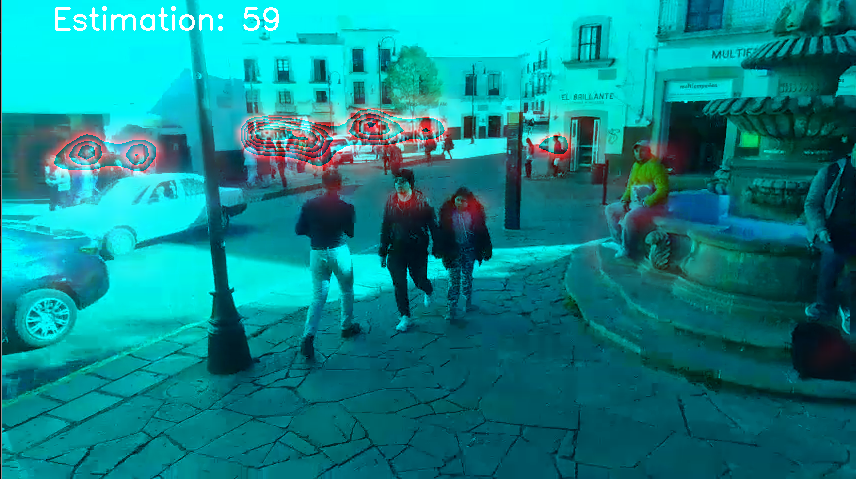}}
\caption{Relevant frame in the fountain video using Faster RCNN as the detector. The estimated count is 28 persons, equal to the ground truth count.}
\label{resnet_fuente}
\end{figure}
\begin{figure}[t]
\centerline{\includegraphics[width=0.5\textwidth]{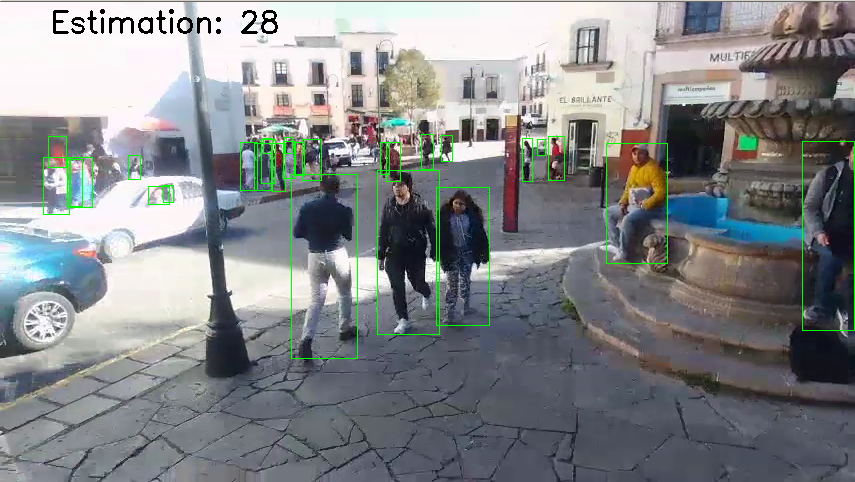}}
\caption{Relevant frame in the fountain video using VGG19 as the detector. The estimated count is 59 persons against the ground truth count of 33 persons. }
\label{vgg_fuente}
\end{figure}
As we can see in Figure \ref{resnet_fuente}, in spite of being close to the ground truth count, Faster RCNN classifies more than once at least 7 persons, increasing the estimated count. This only happens when the pixel representation of the persons is low, or because the person is near the stands that could be identified as another person. The same happened to VGG19 but in grater scale, since the estimated count nearly doubled the ground truth of the frame.
\subsection{Discussion}
\begin{table}[b]
    \centering
    \caption{Count comparison between VGG19, Faster RCNN with ResNet50-FPN and Ground Truth}
    \begin{tabular}{|c|c|c|c|}
    \hline
    \textbf{Video} & \textbf{VGG19} & \textbf{Faster RCNN} & \textbf{Ground Truth} \\
    \hline
    \textit{Garden} & \bf{27} & 14 & 25 \\
    \hline
    \textit{Small square} & 40 & \bf{11} & 17 \\
    \hline
    \textit{Large public square} & \bf{48} & 27 & 49 \\
    \hline
    \textit{Public university} & \bf{19} & 9 & 20 \\
    \hline
    \textit{Fountain}& 59 & \bf{28} & 33 \\
    \hline
    \end{tabular}

    \label{tab:comparision}
\end{table}
VGG19 tended to make big mistakes most notably when rich repetitive patterns and crowds showed together, like in Figure \ref{vgg_fuente}. Nonetheless, it was able to detect almost all the persons that showed in the scene. In dense crowds it was the better approach and had a better performance in therms of frames per second than Faster RCNN.
\\On the other hand, Faster RCNN, in therms of counting estimate, was the most robust when rich patterns appear in the scene, but it tended to miss crowds that were far from the camera and. Moreover, in some cases, it may count more than once each person, and presents more false positive detections. Hence, it is only suitable for sparse crowds near the camera.

\section{Conclusions}
\label{sec-conclusions}
In this paper, we presented two approaches to perform the crowd  detecting and counting tasks using a moving camera mounted on a drone. We conclude that VGG19 and density maps had the best results in both tasks, since it does not depended on the scale or scene context. Meanwhile, Faster RCNN could be utilised for low crowded scenes where a powerful ground computer is available, and rich repetitive patterns are expected. In other words, Faster RCNN may offer a suitable alternative in scenes with sparse crowds with people near the camera.\\
In future work, we could develop a new and lightweight architecture using the Bayes loss function to be able to perform these tasks in real-time, embedded on the drone. Also, it would be interesting to create a new database including rich repetitive patterns in sparse crowds instances, in order to overcome the problems discussed in the paper.

\printbibliography

\end{document}